\documentclass[final,3p,authoryear,11pt]{elsarticle}

\usepackage{amsmath}
\usepackage{amssymb}
\usepackage{endnotes}
\usepackage{lineno}
\usepackage{url}
\usepackage{booktabs}
\usepackage{siunitx} 
\usepackage{xcolor}
\usepackage{multirow}

\usepackage{color}
\definecolor{RoyalBlue}{cmyk}{1, 0.50, 0, 0}

\usepackage[colorlinks]{hyperref}
\AtBeginDocument{%
  \hypersetup{
    allbordercolors={1 1 1},
    allcolors={RoyalBlue}}}

\journal{ArXiv}

\begin{document}
\begin{frontmatter}

\indent \textcolor{RoyalBlue}{Cite as: Garcia, G.P.B., Soares, L.P., Espadoto, M., Grohmann, C.H. 2023. Relict landslide detection using deep-learning architectures for image segmentation in rainforest areas: a new framework. \textit{International Journal of Remote Sensing}, 44(7): 2168–2195. doi:\href{http://dx.doi.org/10.1080/01431161.2023.2197130}{10.1080/01431161.2023.2197130}} \\

\vspace{15pt} 

\title{Relict landslide detection using Deep-Learning architectures for image segmentation in rainforest areas: A new framework}

\author[igc,spam]{Guilherme P.B. Garcia}
\ead{guilherme.pereira.garcia@usp.br}

\author[igc,spam]{Lucas Pedrosa Soares}
\ead{lpsoares@usp.br}

\author[ime]{Mateus Espadoto}
\ead{mespadoto@gmail.com}

\author[iee,spam]{Carlos H. Grohmann\corref{cor1}}
\ead{guano@usp.br}

\cortext[cor1]{Corresponding author}

\address[iee]{Institute of Energy and Environment, University of S\~{a}o Paulo (IEE-USP), S\~{a}o Paulo, 05508-010, Brazil}

\address[igc]{Institute of Geosciences, University of S\~{a}o Paulo (IGc-USP), S\~{a}o Paulo, 05508-080, Brazil}

\address[ime]{Institute of Mathematics and Statistics, University of S\~{a}o Paulo (IGc-USP), S\~{a}o Paulo, 05508-090, Brazil}

\address[spam]{Spatial Analysis and Modelling Lab (SPAMLab, IEE-USP)}

\begin{abstract}
Landslides are destructive and recurrent natural disasters on steep slopes and represent a risk to lives and properties. Knowledge of relict landslides' location is vital to understand their mechanisms, update inventory maps and improve risk assessment. However, relict landslide mapping is complex in tropical regions covered with rainforest vegetation. A new CNN framework is proposed for semi-automatic detection of relict landslides, which uses a dataset generated by a k-means clustering algorithm and has a pre-training step. The weights computed in the pre-training are used to fine-tune the CNN training process. A comparison between the proposed and the standard framework is performed using CBERS-04A WPM images. Three CNNs for semantic segmentation are used (Unet, FPN, Linknet) with two augmented datasets. A total of 42 combinations of CNNs are tested. Values of precision and recall were very similar between the combinations tested. Recall was higher than 75\% for every combination, but precision values were usually smaller than 20\%. False positives (FP) samples were addressed as the cause for these low precision values. Predictions of the proposed framework were more accurate and correctly detected more landslides. This work demonstrates that there are limitations for detecting relict landslides in areas covered with rainforest, mainly related to similarities between the spectral response of pastures and deforested areas with \textit{Gleichenella sp.} ferns, commonly used as an indicator of landslide scars. 
\end{abstract}


\end{frontmatter}



\section{Introduction}
Landslides are destructive and recurrent natural disasters that represent a risk to lives and properties when near urban areas \citep{Metternicht2005,Kasai2009,Netto2013,Jebur2014,Nohani2019}). They are responsible for expressive human and economic losses worldwide, costing millions of dollars each year \citep{tominaga2009desastres,Netto2013}. The Sendai framework for disaster risk reduction 2015-2030 \citep{UNISDR2015} stated that natural hazards affected more than 25 million people and caused economic losses up to US\$ 1.3 trillion between 2008 and 2012. Landslides occur on steep slopes and are essential agents in landscape evolution by promoting changes through successive events that shape the hillsides \citep{Wolle1988,Summerfield1991,tominaga2009desastres,Guzzetti2012}. Deforestation, inadequate urban growth, and climate change are increasing the occurence of mass movements, mainly water-related landslides such as earthflows and mudflows \citep{tominaga2009desastres, Gariano2016,Nohani2019}.

In recent years, ongoing technological development provided new tools for researchers which are faster, better, and more accurate than conventional ones, making data easier to gather and handle which significantly changed the methods of landslide studies \citep{Mantovani1996,Metternicht2005,Scaioni2014}.
Conventional methods consist of extensive fieldwork, scanning of topographic maps and visual photo-interpretation of stereo images, which are costly, time consuming and have limitations that affect the quality of the data \citep{Nilsen1973,Guzzetti1999,Roering2005,VanDenEeckhaut2005,Booth2009,Burns2009,Guzzetti2012,Roering2013,Scaioni2014}. Use of lidar (Light Detection and Ranging) \citep{McKean2004,Glenn2006,Ardizzone2007,VanDenEeckhaut2007,Baldo2009,Burns2009,Kasai2009,Ventura2011,Jaboyedoff2012,Chen2013,Wang2013,Jebur2014}, RPA (Remotely Piloted Aircraft) SfM-MVS (Structure from Motion Multi-view stereo) \citep{Niethammer2010,Lucieer2014,Turner2015,Lindner2016,Yu2017,Mozas-Calvache2017,Menegoni2020,Devoto2020,Samodra2020,Xu2020,Godone2020} and process automation \citep{VanDenEeckhaut2007,Guzzetti2012,VanDenEeckhaut2012,Scaioni2014,Knevels2019} are the hot spot in landslide studies that are replacing the traditional methods quickly. High resolution (HR) and very high resolution (VHR) data from both remote sensing imagery (satellite, RPA) and topographic data (lidar, SfM-MVS) became mandatory tools in landscape studies \citep{Metternicht2005,Jaboyedoff2012,Scaioni2014}).

Satellite multi-spectral imagery and airborne photographs are generally collected soon after the landslide events, which makes visual identification easier due to removal of material and vegetation \citep{Guzzetti1999,Du2007,Booth2009,liu2009geomorphometry,Burns2010,Sameen2019}. Landslide mapping enables the generation and update of inventory maps that are essential input data for risk assessment studies and landslide prediction \citep{Sameen2019,Yu2021,Dias2021}. The knowledge of the exact landslide location allows specific analysis for emergency response and precautions actions\citep{Chen2018}, so a continuous monitoring of landslides and update of inventory maps is recommended. Nonetheless, these are not practicable for many government agencies due to lack of resources, personal or technology, where inventory maps do not exist or those that do exist are outdated \citep{Dias2021}. In this case, inventory maps can be created or updated by identification of relict landslides, but only those which have been preserved in landscape since the day of occurrence.

Identification of relict landslides is important to enhance understanding of landslide causative factors and mechanisms \citep{Sameen2019, Li2021}, to predict future events, design preventive frameworks \citep{Schulz2007} and for quick landslides hazard emergency response \citep{Chen2018}. Also, landslide susceptibility is higher in surrounding areas of relict landslides due to terrain conditions that makes the slope more likely to landslides occurence \citep{Shahabi2015, Zhong2019}. Relict landslide detection is most needed where there is a lack of data, no historical records or outdated inventory maps, such as in many states of Brazil \citep{Dias2021}. However, identification of relict landslides by remote sensing data is not an easy task. In natural steep slopes, vegetation growth covers devastated areas making landslide identification and monitoring harder over time \citep{LEHMANN2008,Portela2014,Scaioni2014}. Furthermore, relict landslide may suffer from surface erosion processes, such as runoff \citep{Fiori2015} or swell factor \citep{Dewitte2005,Pedrazzini2010,Schulz2018}, which disturbs the terrain and degrades landslide boundaries making them unrecognizable.

The post-event visual identification is commonly performed by experts in GIS (Geographic Information System) software and is an exhaustive, time-consuming task \citep{VanDenEeckhaut2007,Burns2009,Burns2010}. In the last couple of decades, semi-automatic and automatic detection of landslides started to be exploited from different sources and tools, mainly using very-high-resolution (VHR) topographic data, such as lidar DEMs (Digital Elevation Models), and multispectral imagery \citep{McKean2004,Glenn2006,Ardizzone2007,VanDenEeckhaut2007,Baldo2009,Burns2009,Kasai2009,Ventura2011,Jaboyedoff2012,Chen2013,Wang2013,Jebur2014}. Automatic landslide detection is important because it allows rapid mapping procedures with potential applications for hazard assessments, risk mitigation, and post-event recovery efforts \citep{Guzzetti2012}. Methods such as OBIA (Object-based Identification Analysis) \citep{Petropoulos2012,VanDenEeckhaut2012,Scaioni2014,Knevels2019}, pixel-based classification \citep{Li2020, Wang2020}, machine learning algorithms \citep{Pal2006,Maxwell2018, Ghorbanzadeh2019, Zhong2019} and Deep Learning \citep{Luus2015,Ding2016,Scott2017,Kussul2017,Chen2018,Ma2019,Sameen2019,Ji2020,Li2020,Prakash2020,Li2021,Yu2021,Soares2022,Xu2022,Meena2022} are the most prominent techniques in landslide semi-automatic and automatic detection studies. 

Machine Learning algorithms and, more recently, Deep Learning neural networks showed excellent performance in remote sensing analysis, although deep networks were not designed to process high-resolution images \citep{Audebert2016,Cheng2017}. Convolutional Neural Networks (CNN) is probably the most successful network architecture in deep learning and have been widely used to extract spatial features for object detection and image segmentation of high-resolution images \citep{Krizhevsky2012,Castelluccio2015,Zhang2016,Kussul2017,Zhao2017,Cheng2018,Ma2019,Zhong2019,Hoeser2020,Li2020,Yu2021,Bai2022}. 

Studies for automatic landslide detection using CNN are still incipient, but the results are promising, and the interest of geoscientists in these techniques is increasing significantly \citep{Ji2020}. Although landslide detection studies share the same final goal, the methods and data used are quite different. Studies for comparison, evaluation of existing  machine learning methods \citep{Sameen2019,Ghorbanzadeh2019,Prakash2020,Wang2020,Li2021} and proposal of new CNN methods \citep{Ding2016,Ji2020,Yu2021,Soares2022} are the most outstanding. Usually, new CNN models are proposed to solve specific problems that standard CNN models had difficulties with or achieved inappropriate results. Introduction to new networks architecture, modification of CNN framework, pre-processing of input data and image post-processing steps are the most common topics of the CNN models proposal studies.

Input data such as multispectral imagery, topographical data, or a combination of both is selected according to data availability and the study area characteristics. Also, the amount of data used for CNN training is essential for obtaining satisfactory results, mainly labeled data \citep{Ronneberger2015}. Most landslide detection studies focus on identifying recently triggered landslides using mainly multispectral images collected soon after the occurrence when landslide are more clearly visible in the landscape, which facilitates the classification process by the CNN. 

On the other hand, identification of relict landslides has difficulties, such as preservation of landslide boundaries and degree of vegetation cover, and is usually performed using as input a DTM (Digital Terrain Model) generated from VHR topographical data that is able to filter out the vegetation cover \citep{Guth2021,Wang2020}. However, in addition to VHR topographical data not existing or not being available for most countries, results from landslide detection studies that added topographical data as an input band in the CNN's training process did not achieved better results than using only spectral bands \citep{Ghorbanzadeh2019,Sameen2019,Ji2020,Soares2022}. 

The main objective of this paper is to perform a semi-automatic detection of relict landslide using CNN for an area in southeastern Brazil that has preserved landslides on its slopes for several decades due to an specific fern species. A new framework for CNN semantic segmentation using transfer learning is proposed as an attempt to overcome difficulties such as lack of labeled landslide data, and improve the model's classification ability. A comparison between the proposed and the standard framework will be performed. The focus is on taking advantage of the visual contrast between the ferns and the native vegetation to perform the detection of relict landslides using CBERS-04A WPM multispectral images as input to the CNN training process. Detection of relict landslides will help updating the inventory map of the region that is dated from the 1970s and did not suffer great reviews \citep{Correa2017,Dias2021}. This study is novel in terms of using CBERS4A multispectral images as input for semi-automatic relict landslide detection and for proposing a new framework for CNN semantic segmentation that performs a pre-training step which is later used to fine tune the neural network, compared to previous studies that attempted to detect landslides in tropical regions.

\section{Methods}

\subsection{Study Area}

\begin{figure}[!ht]
    \begin{center}
    \includegraphics[width=0.8\columnwidth,angle=0]{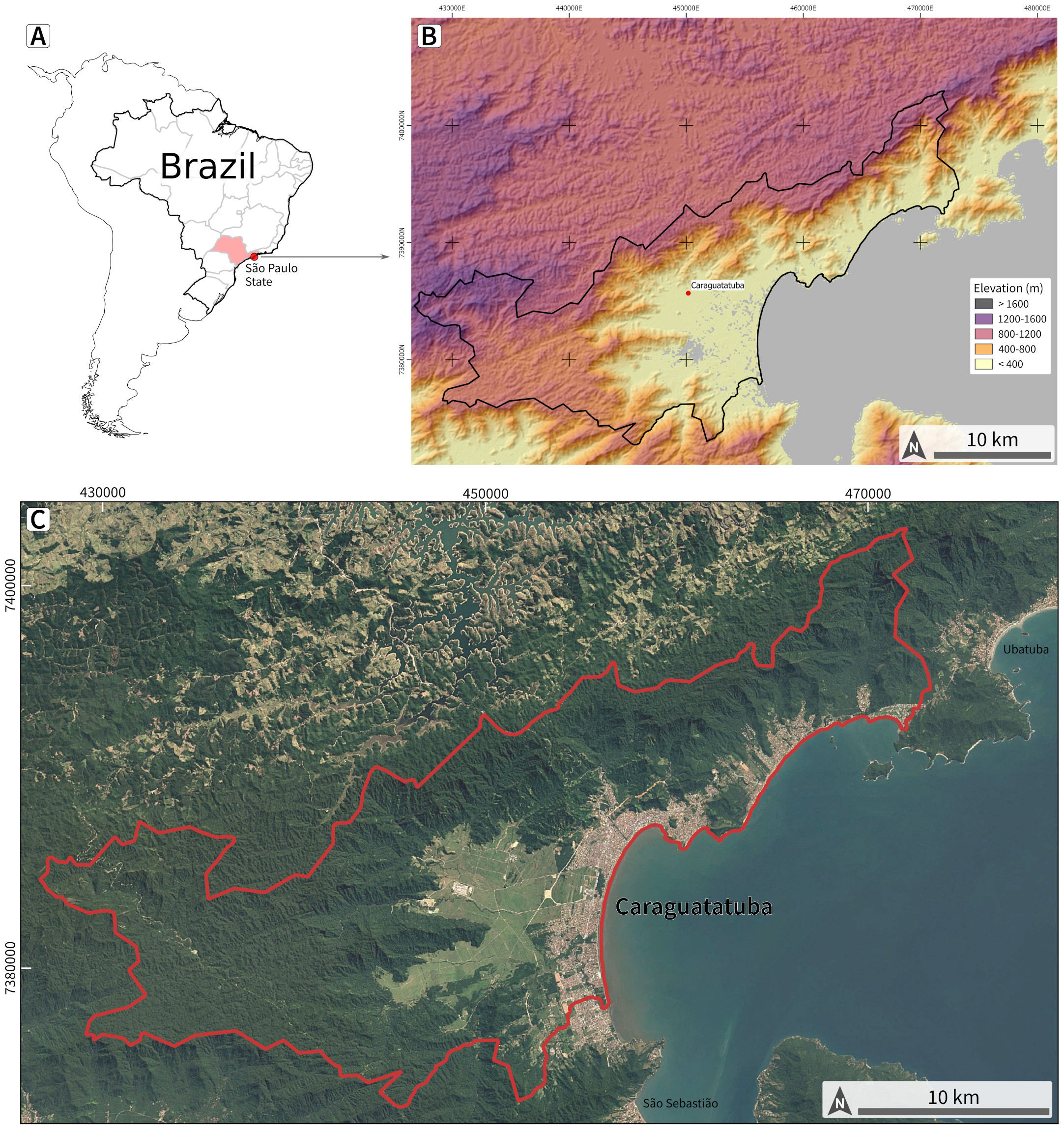}
    \caption[Study area.]{A) Location of S\~{a}o Paulo state in Brazil. B) Elevation map of the study area. C) Image of the study area with Caraguatatuba municipality in red. Satellite image Landsat/Copernicus (2015-12-30), powered by Google. UTM coordinates, zone 23 (South), WGS84. } 
    \label{fig:Studyarea}
    \end{center}
\end{figure}

The study area is within the Serra do Mar mountain range, southeastern Brazil (Fig.~\ref{fig:Studyarea}). It includes the majority of Caraguatatuba municipality, encompassing coastal and mountainous portions. This region is known for the preserved Atlantic rainforest and has a long history of landslides occurrences in the hillsides, mainly shallow landslides and flows \citep{fulfaro1976escorregamentos,nieble1984estabilidade,augusto1992caracterizaccao}. An inventory map is available for a major event that occurred in 1967, when almost 600 landslides of various sizes were identified, which caused destruction and human losses \citep{fulfaro1976escorregamentos}. Despite the high landslide occurrence rate in the region, there is a lack of landslide documentation, with few studies updating the previously inventory map and monitoring the landslides \citep{Correa2017, Dias2021}.

\begin{figure}
    \begin{center}
    \includegraphics[width=0.98\columnwidth,angle=0]{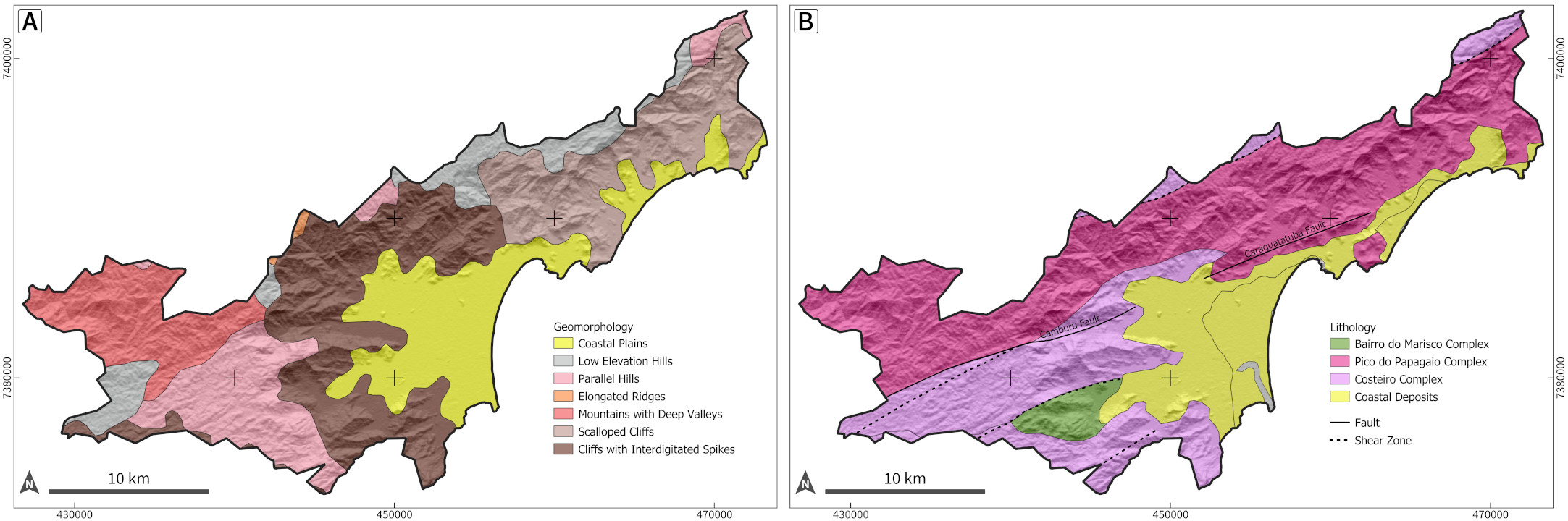}
    \caption{A) Geological and (B) geomorphological map of the study area. UTM coordinates, zone 23 (South), WGS84.} 
    \label{fig:Lito_map}
    \end{center}
\end{figure}

A mountainous landscape marks the study area in the west and the coastal areas with sand beaches and plains in the east \citep{ponccano1981mapa} (Fig.~\ref{fig:Lito_map}A). The mountains are characterized as rugged relief with high slope gradients and valleys with amplitudes higher than 100 m, defined as Costeira Province \citep{Almeida1964,ponccano1981mapa}. The landslides usually occur in this section due to its favorable conditions for mass movements.

The geological setting is within the Mantiqueira Province, specifically in Serra do Mar Domain \citep{Perrota2005}, mainly composed of granites and gneisses from the Costeiro Complex and Pico do Papagaio Complex, which occurs as intercalated lenses controlled by shear zones (Fig.~\ref{fig:Lito_map}B). Several faults and shear zones oriented towards SW-NE comprising the structural framework of the study area \citep{Perrota2005}. The principal structure is the shear zone that occurs in the N-NW portion of the study area, namely the Bairro Alto Shear zone; other important structures are the Caraguatatuba fault and the Camburu fault. These structures control lithology. There is also the presence of mafic intrusions related to the Ara\c{c}ua\'{i} - Rio Doce orogen, quaternary deposits occur in river banks and coastal plans.

The local climate also contributes to landslide occurrence with rainy summers that usually trigger landslides and other mass movements \citep{ContiFurlan19998}. Climate regime in Caraguatatuba is defined as Cwa, or altitude tropical climate \cite{koeppen1948climatologia}. The annual average rainfall is up to 1830 mm, with August as the driest month (60 mm) and January as the rainiest (300 mm).

Most of the study area is within the Atlantic rainforest biome, which covers almost all Brazilian coastal region and presents great biodiversity. The Dense Ombrofile Forest is the predominant vegetation cover, distinguished by high indices of temperature and rainfall during the year \citep{ellenberg1967key,veloso1991classificaccao}, evergreen vegetation cover with several layers, a canopy that can reach up to 50 m in height, and dense shrubbery detaching ferns, bromeliads, and palm trees \citep{Portela2014}. Forest formation is controlled by elevation due to its mountainous context and can be divided into Dense Ombrofile Forest, Arboreal Forest, and Secondary vegetation \citep{kronka2007inventario}.

Despite the preserved vegetation in most of the study area, the landslides occur on natural steep slopes but are more common in deforested areas and close to roads and urban areas.
\subsubsection{\textit{Gleichenella sp.}}
\label{sec:gleichenella}

\begin{figure}[!ht]
    \begin{center}
    \includegraphics[width=0.6\columnwidth,angle=0]{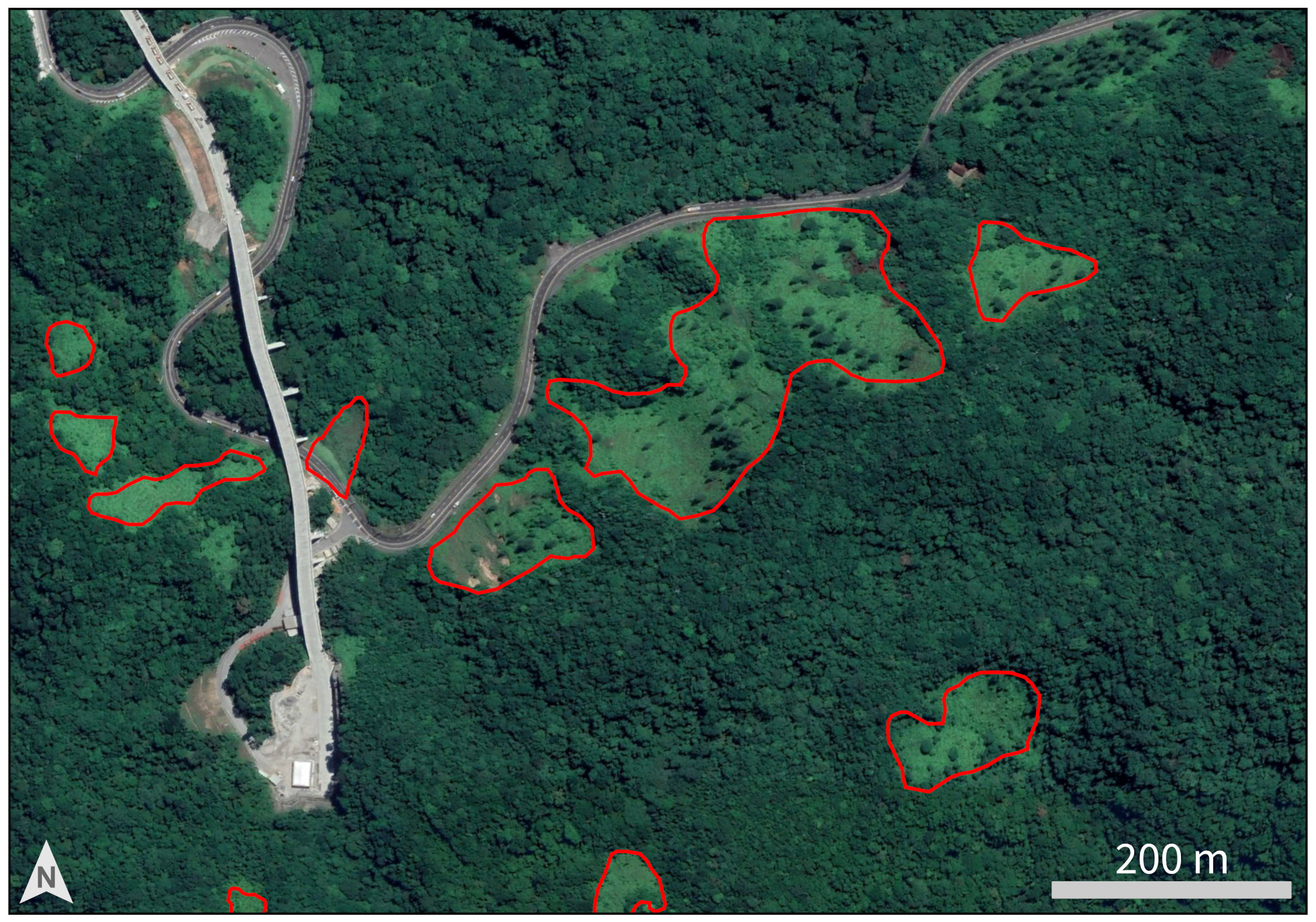}
    \caption[Gleichenella]{Example of \textit{Gleichenella} ferns occurrence in landslide scars in Caraguatatuba municipality. Satellite imagery \textcopyright2021 Digital Globe, powered by Google. See location in Figure~\ref{fig:Studyarea}} 
    \label{fig:Gleichenella}
    \end{center}
\end{figure}

Ferns are common plants in the Atlantic rainforest of Serra do Mar being pioneers and efficient in regenerating degraded forest \citep{LEHMANN2008}. The hillsides degraded by landslides in the study area are usually covered by a specific fern species named \textit{Gleichenella sp.} while others are almost totally recovered by the forest \citep{LEHMANN2008,Portela2014} (Fig.~\ref{fig:Gleichenella}). \textit{Gleichenella sp.} usually prevents complete forest regeneration in the degraded areas, maintaining the landslide scar and boundaries distinguishable from adjacent areas.

\subsection{Data}

\subsubsection{CBERS-04A}

CBERS-04A is a mid-resolution remote sensing satellite in a sun-synchronous orbit launched in December 2019, with the first images released for download in January 2020. The satellite is operated by the Brazilian National Institute of Spatial Research (INPE) and the images are available for free download on their website (http://www2.dgi.inpe.br/catalogo/explore). It is equipped with three cameras: MUX (Multispectral Camera), WFI (Wide Field Imager), and WPM (Wide Panchromatic Multispectral). MUX and WFI have four optical bands (R, G, B, and NIR) with spatial resolution of 16.5 m and 55.5 m, respectively. WPM has five bands (R, G, B, NIR, and Panchromatic) with spatial resolution of 8 m for the optical bands and 2 m for the panchromatic. The revisit period is 31 days for the WPM and MUX and five days for the WFI. The swath width for each camera is 92 km (WPM), 95 km (MUX), and 684 km (WFI).

In this project five CBERS4A images were used with the four optical bands of the WPM camera, namely blue (0,45 - 0,52 $\mu$m), green (0,52 - 0,59 $\mu$m), red (0,63 - 0,69 $\mu$m) and near-infrared (0,77 - 0,89 $\mu$m). All the images were used without radiometric or atmospheric corrections and with no cloud cover.

\subsection{Convolutional Neural Networks}

CNN has shown impressive performance in many applications, including remote sensing analysis, with fast growth in the use of this network in the last few years \citep{Krizhevsky2012,Castelluccio2015,Cheng2018,Bai2022}. The typical architecture of a CNN is composed of multiple feature-extraction stages where each stage consists of a series of layers, including convolutional layers, pooling layers, and fully connected layers \citep{Castelluccio2015,Cheng2018}. CNN is designed to take advantage of the two-dimensional structure of the input image and focus on bridging the low-level features to high-level semantics of the image scene, automatically extracting intrinsic features from remote sensing imagery \citep{Zhang2016,Zhong2017}. For \cite{Zhao2015} a limitation for using CNN in remote sensing image classification is that it requires many labeled training samples, which is not always available.

In this study, the focus was not to propose a new CNN architecture, but to explore the capabilities of the existing CNN models in the Segmentation Models python library \citep{Yakubovskiy:2019}. The CNNs in these library are built explicitly for semantic segmentation tasks rather than classification, which means that the classification task aims to assign one label to each pixel in the images \citep{Bai2022}. U-Net \citep{Ronneberger2015}, FPN \citep{Lin2017} and Linknet \citep{Chaurasia2017} were the CNN models chosen for landslide detection in this project (Fig.~\ref{fig:CNNs}).

%


\subsubsection{Deep Learning Architectures for Image Segmentation}

Image segmentation models are considered to be a multi-scale context problem, i.e., to predict the class of a single pixel they use contextual information, so the areas around the pixel of interest help in its identification, based on some factors like the size and continuity of the element, and the amount of neighbouring segments of other classes \citep{Hoeser2020}. Landslide detection studies benefit from contextual information since landslides, particurlaly shallow landslides, have a distinctive color and long, narrow shapes that distinguishes them from their surroundings, both in natural slopes and urban areas, which facilitates the identification through image segmentation \citep{Ghorbanzadeh2019,Sameen2019,Prakash2020,Soares2022}. 

\begin{figure}[!ht]
    \begin{center}
    \includegraphics[width=0.7\columnwidth,angle=0]{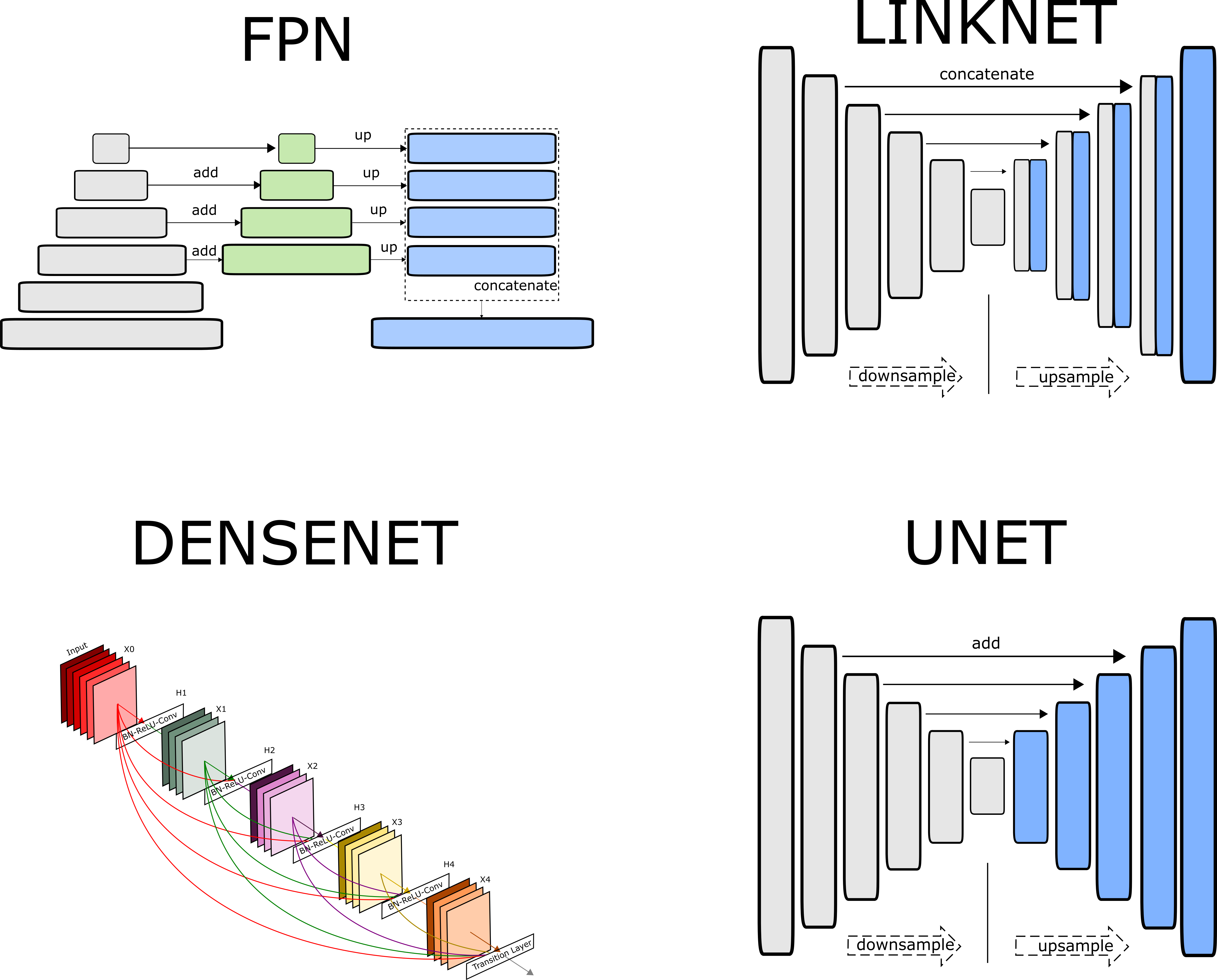}
    \caption[CNNs]{Structure of FPN, LinkNet, DenseNet and Unet. Modified from \cite{Yakubovskiy:2019}} 
    \label{fig:CNNs}
    \end{center}
\end{figure}

The Unet architecture was introduced by \cite{Ronneberger2015}, modifying and extending FCN (Fully Convolutional Network) by working with very few training images, with more accurate segmentation and preserving image localization. The main difference from traditional CNN lies in its architecture of the expanding path. Unet consists of two paths: a constructive path and an expansive path (Fig.~\ref{fig:CNNs}). The constructive path follows the typical CNN architecture with convolution and pooling layers for downsampling. The expansive path consists of upsampling the feature map and replacing the fully connected layers. 

Feature pyramids are a primary component in recognition systems for detecting objects at different scales. The Feature Pyramid Network (FPN) was introduced by \citet{Lin2017} as a feature pyramid with rich semantics at all levels, is built quickly from a single input image scale, and is a generic solution built inside a deep CNN. Their architecture combines low-resolution, semantically strong features, with high-resolution, semantically weak features, by a top-down pathway and lateral connections (Fig.~\ref{fig:CNNs}). The pyramid's construction involves two parts: a bottom-up pathway and a top-down pathway, and the lateral connections between them.

LinkNet is a neural network architecture proposed by \citet{Chaurasia2017} designed specifically for semantic segmentation (Fig.~\ref{fig:CNNs}). It comprises an encoder-decoder pair containing residual blocks and linking each encoder with a decoder to avoid losing spatial information. The input of each encoder layer is bypassed by the output of its corresponding decoder so that the decoder and the upsampling operations can use the spatial information. LinkNet results were more efficient compared to existing state-of-the-art segmentation networks that have an order of magnitude larger computational and memory requirements.

DenseNet \citep{Huang2017} was used as a backbone for the pre-training step and it is available as standard backbone in the segmentation models python library \citep{Yakubovskiy:2019}.  It was proposed as a network with a different connectivity pattern that connects each layer to every other layer in a feed-forward fashion, namely Dense Convolutional Network (DenseNet) (Fig.~\ref{fig:CNNs}). A remarkable difference between DenseNet and other networks, such as ResNet \citep{He2016}, is that it yields condensed models by feature reuse that are easy to train and more efficient. It also concatenates feature maps learned by different layers, which increases variation and efficiency. The DenseNet architecture comprises dense blocks and transition layers, with a convolution at the beginning and a softmax classifier at the end.

\subsubsection{Proposed Deep Learning framework}

\begin{figure}[!ht]
    \begin{center}
    \includegraphics[width=0.7\columnwidth,angle=0]{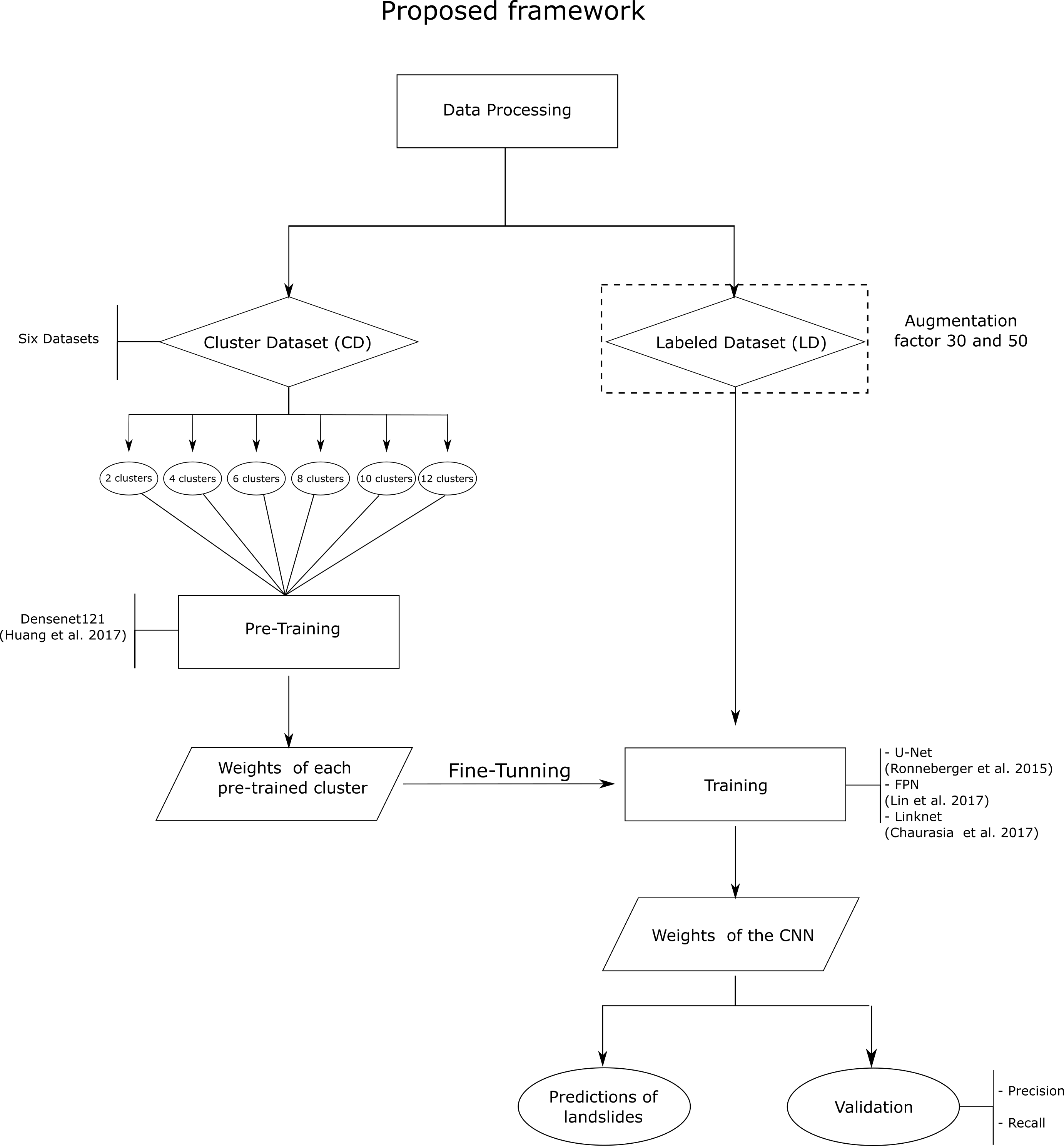}
    \caption[Framework]{Proposed framework of CNN relict landslide detection.} 
    \label{fig:Frameworks}
    \end{center}
\end{figure}

This study proposes a new deep learning framework for semi-automatic relict landslide detection that uses transfer learning for accuracy improvement, and the results are compared to a standard CNN framework. The standard framework is usually performed in three steps: data processing, training, and validation, while the proposed framework has one more step, pre-training (Fig.~\ref{fig:Frameworks}).

\begin{itemize}
  \item Data Processing
    
    
    Two datasets were created in the data processing step, the Labeled dataset and the Cluster Dataset. The Labeled Dataset comprises images labeled in two classes, with or without landslide scars. It uses the CBERS4A 201/143 (path/row) image from August of 2020 as input with a dimension of 14210 x 14592 pixels which was clipped in tiles of 32x32 pixels. A zero-padding process was used to enable the creation of the tiles. Each tile received a label due the existence of landslides (label = 1) or not (label = 0) according to a landslide mask. The landslide mask was created by rasterizing a landslide vector composed by 384 landslides that were identified from \citet{fulfaro1976escorregamentos} inventory map and visual interpretation of current images. If a tile has at least 1 pixel in intersection with the landslide mask, it is labeled as landslide. The CBERS4A 201/143 image that encompasses the study area was split in train/test areas so that the landslide mask was split into a nearly 70/30 ratio, where 276 ($\sim$70\%) landslides belong to the train area and were used to create the labeled dataset (LD) and 108 ($\sim$30\%) belong to the test area and were used for evaluation of model's accuracy (Fig.~\ref{fig:Train}). The Labeled Dataset has a total of 202464 tiles of which only 422 tiles were labeled as landslides (Table~\ref{LD}).

    The Cluster dataset (CD) used the k-means algorithm to cluster data by trying to automatically separate samples into \textit{k} groups based on pixel color similarity which is expected to be able to segregate major land cover areas \citep{scikit-learn}. Images covering the full extension of the Serra do Mar mountain range were used in this step, almost encompassing the whole coast of S\~{a}o Paulo state. Five CBERS4A images were used as input (path/row): 200/142, 200/143, 201/142, 201/143 and 202/143. The images were clipped in tiles with 32 x 32 pixels, also using the zero-padding process for the borders of the images. Each tile receives a label with the predominant class according to the k-means clustering results. Since the number of clusters has to be specified for the k-means algorithm, the clustering was performed with values 2, 4, 6, 8, 10, and 12. To enhance the clustering process, pixels representing ocean and urban coastal areas were removed from input images, so the k-means algorithm used only highland and mountain features. Lastly, a class balancing process was performed with the tiles assigned to each cluster (Table~\ref{CD}).

\begin{figure}[!ht]
    \begin{center}
    \includegraphics[width=0.9\columnwidth,angle=0]{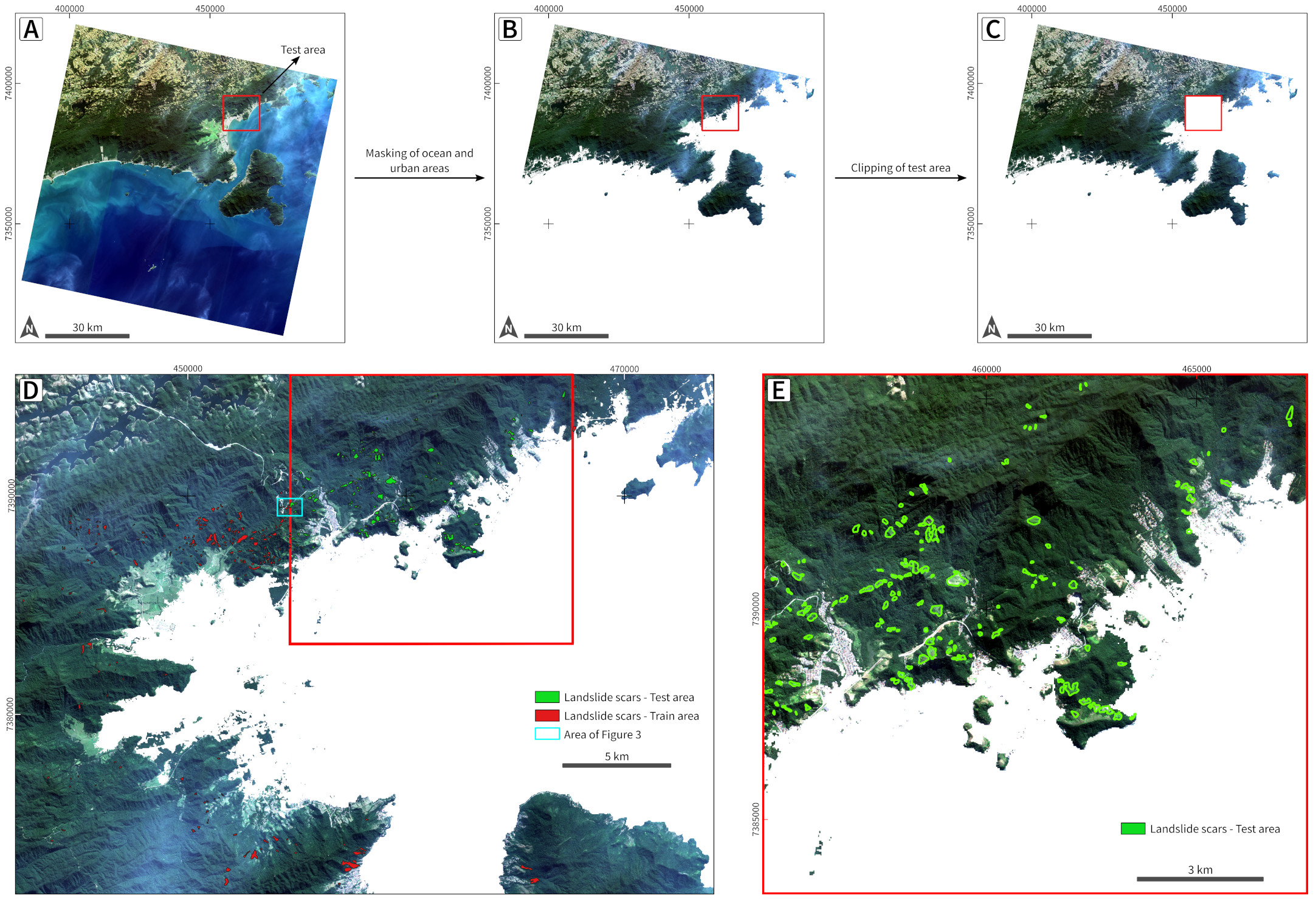}
    \caption[Train]{A) CBERS-04A image 201/143 with the location of the test area. B) Image with ocean and urban areas clipped. C) Image with the test area clipped, used for the training process. D) Landslide scars in the train area and test area. E) Test area with landslide scars.} 
    \label{fig:Train}
    \end{center}
\end{figure}

    \item Augmentation  
    
    Usually, a massive amount of data is necessary for training a neural network such as CNN, which is not always available \citep{Zhao2015}. To overcome this lack of training samples, researchers can use data augmentation techniques to increase these numbers \citep{Ronneberger2015}. This study performed three types of data augmentation techniques: horizontal flip, vertical flip, and a combination of both (horizontal-vertical flip). To avoid unbalancing data, the augmentation was performed only for the positive class of the Labeled Dataset, i.e., tiles labeled as landslide. In total 390 tiles were labeled as landslides and were splitted in train/test sets in a ratio of 60/40, 241 labeled landslides in the train area and 149 labeled landslides in the test area (Table~\ref{LD}). The 241 labeled landslide in the train area were augmented by a factor of 30 (7230 tiles) and 50 (12050 tiles); these augmented datasets are referred to as LD30 and LD50, respectively, and were used in the training process for both the standard and the proposed CNN framework.

  \item Pre Training
    
  Densenet121 was chosen for this pre-training step, and the input was the Cluster dataset, which was explicitly created to pre-train the backbone network. The pre-training step intends to initialize the CNN, i.e., enable the CNN to learn how to segregate objects through the image bands automatically. As spectral features of landslides are usually different from adjacent areas, the pre-training step may help to identify these areas. The weights computed in this process will be used for transfer learning, i.e., fine-tuning the training process of the CNNs. Transfer learning is usually used when there is a lack of training dataset which is case with labeled landslide data for the study area \citep{Liu2018}.

\begin{table}[ht!]
    \centering
    \renewcommand{\arraystretch}{1}
    \begin{tabular}{ccl}
      \multicolumn{3}{c}{\textbf{Labeled Dataset}} \\
      \toprule
       \textbf{Augmentation} & \textbf{Negative Class} & \textbf{Positive class} \\
       \midrule
             & 202,042 tiles & 422 tiles\\

         LD30 & -      & 12,660 tiles \\

         LD50 & -     & 21,100 tiles \\
         \bottomrule
    \end{tabular}
    \caption{Details of Labeled Dataset.}
    \label{LD}
\end{table}

\begin{table}[ht!]
    \centering
    \renewcommand{\arraystretch}{1}
    \begin{tabular}{lllllll}
      \multicolumn{7}{c}{\textbf{Cluster Dataset}} \\
      \toprule
        \textbf{Classes $\downarrow$/Clusters $\rightarrow$} & \multicolumn{1}{c}{\textbf{2}} & \multicolumn{1}{c}{\textbf{4}} & \multicolumn{1}{c}{\textbf{6}} & \multicolumn{1}{c}{\textbf{8}} & \multicolumn{1}{c}{\textbf{10}} & \multicolumn{1}{c}{\textbf{12}} \\
        \midrule
        \multicolumn{1}{c}{\textbf{0}} & 497474 tiles & 165743 & 99447 & 71004 & 55195 & 45147 \\
        \multicolumn{1}{c}{\textbf{1}} & 497474 & 18084 & 5360 & 3713 & 3540 & 3064  \\
        \multicolumn{1}{c}{\textbf{2}} & \multicolumn{1}{c}{-} & 206833 & 20703 & 20277 & 19473 & 17781  \\
        \multicolumn{1}{c}{\textbf{3}} & \multicolumn{1}{c}{-} & 272312 & 259210 & 253449 & 251040 & 13156\\
        \multicolumn{1}{c}{\textbf{4}} & \multicolumn{1}{c}{-} & \multicolumn{1}{c}{-} & 85145 & 35873 & 22875 & 255105\\
        \multicolumn{1}{c}{\textbf{5}} & \multicolumn{1}{c}{-} & \multicolumn{1}{c}{-} & 126818 & 93978 & 39922 & 28676\\
        \multicolumn{1}{c}{\textbf{6}} & \multicolumn{1}{c}{-} & \multicolumn{1}{c}{-} & \multicolumn{1}{c}{-} & 29688 & 25174 & 49795\\
        \multicolumn{1}{c}{\textbf{7}}& \multicolumn{1}{c}{-} & \multicolumn{1}{c}{-} & \multicolumn{1}{c}{-} & 60050 & 58609 & 26133\\
        \multicolumn{1}{c}{\textbf{8}} & \multicolumn{1}{c}{-} & \multicolumn{1}{c}{-} & \multicolumn{1}{c}{-} & \multicolumn{1}{c}{-}  & 29999 & 42012\\
        \multicolumn{1}{c}{\textbf{9}} & \multicolumn{1}{c}{-} & \multicolumn{1}{c}{-} & \multicolumn{1}{c}{-} & \multicolumn{1}{c}{-} & 46127 & 9913\\
        \multicolumn{1}{c}{\textbf{10}} & \multicolumn{1}{c}{-} & \multicolumn{1}{c}{-} & \multicolumn{1}{c}{-} & \multicolumn{1}{c}{-} & \multicolumn{1}{c}{-} & 19052\\
        \multicolumn{1}{c}{\textbf{11}} & \multicolumn{1}{c}{-} & \multicolumn{1}{c}{-} & \multicolumn{1}{c}{-} & \multicolumn{1}{c}{-} & \multicolumn{1}{c}{-} & 31932\\
        \bottomrule
        
    \end{tabular}
    \caption{Detail of cluster dataset with the number of tiles for each class created.}
    \label{CD}
\end{table}

  \item Training
    
    Three CNNs were used for landslide detection: FPN, Linknet, and Unet. These CNNs are available in the Segmentation Models python library based on Keras(Tensorflow) \citep{Yakubovskiy:2019} and are specific for semantic segmentation tasks. All three CNN models have a downsample and an upsample path and use the Densenet121 as a backbone. The weights from these CNNs are evaluated in the test area by validation indices and prediction of landslides. The standard framework used both LD30 and LD50 as input to train each CNN and compute the weights for later validation. A total of 6 combinations were made, and the results were used to predict landslides. In the proposed framework, the weights learned in the pre-training step were used to fine-tune the training process. The CNNs were trained for every cluster with LD30 and LD50 as input, resulting in 12 combinations for each CNN (FPN, Linknet, Unet), 36 in total, used for model evaluation. All CNN training process were made with a learning rate of 0.00001 and for 300 epochs. 

  \item Validation/Accuracy Assessment
    

    Precision and Recall were used to validate the models. These statistical methods are based on three kinds of classified pixels: True Positives (TP), False Positive (FP), and False Negative (FN) \citep{Ghorbanzadeh2019}. Precision (Eq.~\ref{eq:precision}) is used to quantify the rate of positive samples (TP) among predicted positive samples (TP + FP). The higher the precision of the model, the better the probability of correctly classifying positive samples. In other words, precision determines how many of the classified areas are really landslides. Recall (Eq.~\ref{eq:recall}) represents the ratio of correctly predicted landslides to the ground truth \citep{Ji2020}. It means that recall can determine how much of the landslide areas defined by visual interpretation were classified in the images \citep{Ghorbanzadeh2019}.

\end{itemize}

\begin{equation}
 Precision  = \frac{TP}{TP + FP}
 \label{eq:precision}
\end{equation}

\begin{equation}
 Recall  = \frac{TP}{TP + FN}
 \label{eq:recall}
\end{equation}


\section{Results and Discussion} 


The results were evaluated by Precision and Recall validation indices and by landslide predictions that enable visual interpretation. All the combinations of CNNs and parameters were trained and validated in the train area, while the results were obtained from the test area. Table \ref{tbl:PA_Precision_Recall} show the values of precision and recall for every combination tested in the proposed framework. Table \ref{tbl:SA_Precision_Recall} show the values of precision and recall for the standard CNN framework.

\subsection*{Validation Indices}
\label{sec:validation}
The results between the standard and the proposed CNN frameworks are similar, with a difference between the validation results generally around 5\%. The combinations of parameters and clusters also had similar results. Recall achieved relevant outcomes, with values higher than 75\% for every combination tested. U-Net neural network is commonly used for semantic segmentation of landslides in other studies with relevant outcomes \citep{Prakash2020,Li2021,Soares2022}, therefore, it was expected that U-Net combinations would produce the best results in this project, which did not necessarily happen. Although U-Net achieved the best recall and precision results for the standard framework, it was surpassed by FPN and Linknet in the proposed framework. 

The best recall result for the standard framework was 79.52\% (Unet with LD30), while for the proposed framework the best recall result was 81.09\% (Linknet with 4 clusters and LD50). In contrast with recall results, precision values were unexpectedly low for all combinations usually between 10\% and 20\%. The higher values of precision were 15.83\% for Unet with LD50 in the standard framework and 20.47\% of FPN with 8 clusters and LD30 for the proposed framework. These low precision values may indicate that the models have a high false positive rate, i.e., non-landslide pixels classified as landslides (Eq.~\ref{eq:precision}).

In the studies of semi-automatic landslide detection using CNN, the values of recall and precision are both commonly above 70\% \citep{Chen2018,Ding2016,Ghorbanzadeh2019,Ji2020,Li2021,Yu2021,Zhong2019}. These high rates of recall and precision are probably related to the ease of identifying recently occurred landslides trough images collected right after the event when vegetation cover is removed and the soil is exposed. Nonetheless, maximizing both evaluation indices may be challenging as they are inversely correlated, i.e., increases in recall often come at the expense of decreases in precision.

These results (high recall/low precision) mean that the CNNs tested in this study identifieid, or predicted, more relict landslides than actually exist on the ground truth, which generated many false positives samples, thus decreasing precision. High recall/low precision models are considered less conservative, which means that for landslide detection, the model will try to identify as many landslides as possible at the expense of generating many false positive samples.

Although these results seem may flawed (inappropriate or unsatisfactory) at first, they are consistent with the primary objective of detecting relict landslides. Landslide detection studies are valuable in creating and updating inventory maps with the inclusion of new landslides after they occur. Thus, in areas with few landslide historical data or outdated inventory maps, it is a good sign that landslide detection methods are able to predict more landslides than it exist on ground truth. In other words, false positive samples may indicate new relict landslides identified in the study area that were previously undetected.\\

\begin{table}[ht!]
    \centering
    \begin{tabular}{lclll|lll}
             &   &  \multicolumn{3}{c}{\textbf{Precision}} & \multicolumn{3}{c}{\textbf{Recall}} \\
        \cmidrule{3-8}
        \textbf{Dataset} & \textbf{Clusters} &  \textbf{FPN} & \textbf{Unet} & \textbf{Linknet} &  \textbf{FPN} & \textbf{Unet} & \textbf{Linknet}  \\
        \midrule
         \multirow{6}{*}{LD30} & 2  &  0.1144 & 0.1270 & 0.1367 &  0.7618 & 0.7824 & 0.7749 \\
                               & 4  &  0.1755 & 0.0537 & \textbf{0.1902} &  0.7669 & 0.7965 & 0.7828 \\
                               & 6  &  0.1425 & \textbf{0.1875} & 0.1779 &  0.7820 & 0.7914 & 0.7866 \\
                               & 8  &  \textbf{0.2047} & 0.1644 & 0.0555 &  0.7710 & 0.7944 & 0.7795 \\
                               & 10 &  0.0687 & 0.1655 & 0.1732 &  0.7602 & 0.7956 & 0.8056 \\
                               & 12 &  0.1328 & 0.1676 & 0.1674 &  0.7696 & 0.7775 & 0.7815 \\
         \midrule
         \multirow{6}{*}{LD50} &2  &  0.0473 & 0.1431 & 0.1366 &  0.7581 & 0.7782 & 0.7828 \\
                               & 4  &  0.0876 & 0.0621 & 0.1752 &  0.7788 & 0.7886 & \textbf{0.8109} \\
                               & 6  &  0.1542 & 0.1585 & 0.1546 &  0.7765 & 0.7798 & 0.7832 \\
                               & 8  &  0.1836 & 0.1428 & 0.0612 &  0.7750 & 0.7800 & 0.7996 \\
                               & 10 &  0.1786 & 0.0627 & 0.0471 &  \textbf{0.7853} & \textbf{0.8095} & 0.7906 \\
                               & 12 &  0.1323 & 0.1723 & 0.1613 &  0.7692 & 0.7877 & 0.7970 \\
         \bottomrule
    \end{tabular}
    \caption{Precision and Recall results of FPN, Unet and Linknet models tested for the proposed framework. In bold, the highest value of each model.}
    \label{tbl:PA_Precision_Recall}
\end{table}

\begin{table}[ht!]
    \centering
    \begin{tabular}{llll|lll}
             &   \multicolumn{3}{c}{\textbf{Precision}} & \multicolumn{3}{c}{\textbf{Recall}} \\
        \cmidrule{3-7}
        \textbf{Dataset} &  \textbf{FPN} & \textbf{Unet} & \textbf{Linknet} &  \textbf{FPN} & \textbf{Unet} & \textbf{Linknet}  \\
        \midrule
         LD30  &  \textbf{0.1337} & 0.1405 & 0.1212 &  \textbf{0.7845} & \textbf{0.7952} & 0.7774 \\
         \midrule
         LD50  &  0.1264 & \textbf{0.1583} & \textbf{0.1240} &  0.7750 & 0.7925 & \textbf{0.7892} \\

         \bottomrule
    \end{tabular}
    \caption{Precision and Recall results of FPN, Unet and Linknet models tested for the standard framework. In bold, the highest value of each model.}
    \label{tbl:SA_Precision_Recall}
\end{table}

\subsection*{Predictions}
\label{sec:prediction}

The landslides prediction maps enable visually evaluating the CNNs ability to detect landslides. All the predictions were performed with a threshold of 0.5, i.e., cell values above 0.5 are classified as landslide. Predictions were very heterogeneous across all tested combinations, with some models correctly detecting relict landslides, while predictions from other models were very noisy, with many errors and unable to detect any relict landslides. Predictions that correctly detected most relict landslides did so with errors and misclassifications, mostly false positive samples. 

\begin{figure}[!ht]
    \begin{center}
    \includegraphics[width=0.9\columnwidth,angle=0]{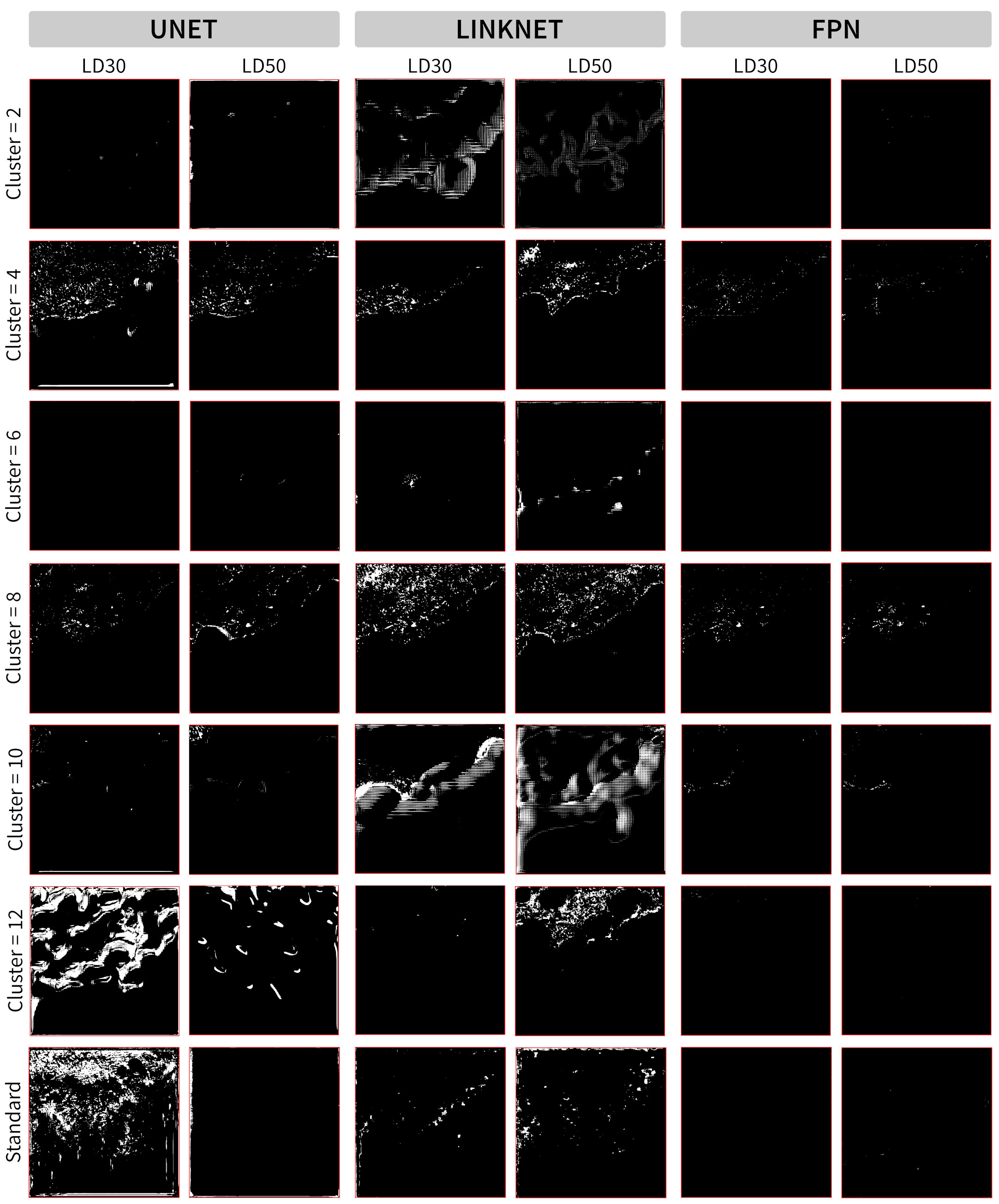}
    \caption[Predictions]{Predictions of landslides from the proposed and standard frameworks.} 
    \label{fig:Predictions}
    \end{center}
\end{figure}

Although recall assessment showed high results for all the combinations from both frameworks, mainly above 75\%, predictions showed that only 18 of 42 (36 + 6) combinations were able to detect relict landslides more or less accurately (Fig.~\ref{fig:Predictions}). Most of the predictions did not detected any relict landslide or had a coarse aspect with many false positive samples.

\begin{figure}[!ht]
    \begin{center}
    \includegraphics[width=0.8\columnwidth,angle=0]{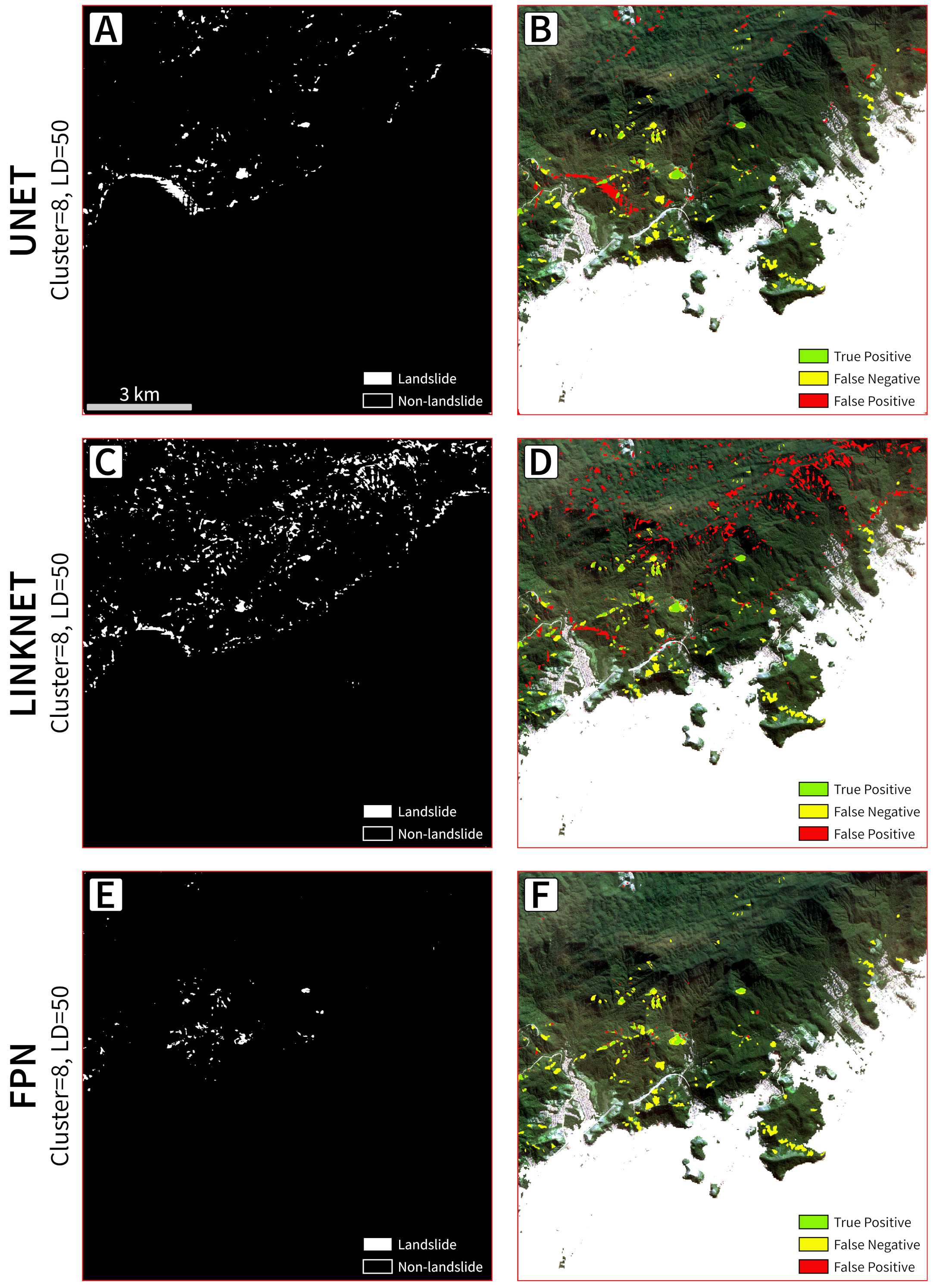}
    \caption[TP]{Image of True Positive, False Positive and False Negative samples from the predicitons of (A) Unet, (B) Linknet and (C) FPN. All predictions are from the combination of 8 clsuters and LD50.} 
    \label{fig:TP}
    \end{center}
\end{figure}

Predictions of the proposed framework were more acurrate than those of the standard framework. 15 of the 36 (41.7\%) predictions of the proposed framework detected relict landslide in the test area, but always with the presence of false positive samples. Figure~\ref{fig:TP} shows predictions of the CNNs with 8 clusters and LD50, as well as TP, FP and FN results. The supplementary material presents all 15 predictions that correctly detected landslides with TP, FP and FN results.

In general, these predictions managed to delineate the relict landslides mainly located in the center-north portion of the test area. Predictions of Unet (6 clusters and LD50) and Linknet (6 clusters with LD30 and LD50) identified only the relict landslides in the center-south portion (Fig.~\ref{fig:Predictions}).

It is not clear why this difference in classification per regions occurs. The southern portion of the study area represents mainly urban and ocean areas that were removed from the original image in the data processing step, and the voids were labeled as 'no data.' The image tiles used as input for the CNN were generated from this clipped image. The tile size (32 x 32 pixels) and the tile cutting pattern may have affected the accuracy for relict landslide detection of the CNN. That may help explain why the landslides in the southern portion were more complex to detect than those from the center-north portion of the study area.

For the standard framework 3 of 6 (50\%) predictions (Linknet with LD30/LD50 and Unet with LD30) correctly detected relict landslides, but the latter also showed a high number of false positives and had a noisy representation similar to salt-and-pepper effect. However, despite being less accurate than the proposed framework predictions, these predictions were able to detect relict landslides in the whole study area.

Notably, there is a clear difference in the number of false positive samples in the landslide predictions of each method. FPN predictions are those with fewer false positive samples, which means that FPN is the most conservative method, but also with fewer true positive samples. The best FPN predictions are those with 8 clusters and both LD30 and LD50, that can detect almost all the major relict landslides in the area with very few errors. 

U-Net and Linknet predictions are less conservative models, i.e., predicted many false positive samples, in an attempt to detect more relict landslide than it exist on the ground truth. The problem in the less conservative models is that they are subject to many errors, some of those very rude. There are some false positive samples in standardized, rectilinear or symmetrical positions, which are very strange and unatural positions. The combinations of (CNN/Cluster/LD) Linknet/4/LD50, Linknet/8/LD50, Linknet/Standard/LD30, Linknet/Standard/LD50, U-Net/4/LD30, U-Net/4/LD50 and U-Net/8/LD50 depict these kind of errors. But the most outstanding errors occur with the U-Net/Standard/LD30 predictions which are smeared with false positive samples across the entire test area.\\

\subsection*{Correlation}
\label{sec:correlation}

Although 42 combinations of parameters were performed it was not possible to determine an optimal combination for relict landslide detection. Type of CNN (U-Net, FPN, Linknet), augmented dataset (LD30 or LD50) and number of clusters (2, 4, 6, 8, 10, 12) did not show a consistent and significant correlation with the results. Recall and precision results were very similar for all 42 combinations whereas the predictions were very divergent. The combinations that detected relict landslides in the predictions were not necessarily the same combinations that achieved the best results of recall and precision (Table~\ref{tbl:Predic_recall_pre}). 

The combinations of FPN and Linknet parameters that achieved better recall and precision results were able to detect relict landslides in the predicitons, unlike what happens with the combinations of U-Net parameters where the predictions that detected relict landslide are not the same than those with better recall and precision results. Thus, it demonstrates that one cannot rely only on validation results to evaluate the CNNs, but that visual inspection of landslide predictions is also of paramount importance.

\begin{table}[ht!]
    \centering
    \begin{tabular}{lclll}
        \textbf{Type of CNN} & \textbf{Clusters} & \textbf{Dataset} & \textbf{Recall} & \textbf{Precision} \\
        \toprule
        \multirow{6}{*}{FPN} & \multirow{2}{*}{4} & LD30 & 0.7669 & 0.1755 \\ 
          &  & LD50 & 0.7788 & 0.0876 \\ \cmidrule{2-5}
          & \multirow{2}{*}{8} & LD30 & 0.7710 & \textbf{0.2047} \\ 
          &  & LD50 & 0.7750 & 0.1836 \\ \cmidrule{2-5}
          & \multirow{2}{*}{10} & LD30 & 0.7602 & 0.0687 \\
          &  & LD50 & \textbf{0.7853} & 0.1786 \\
             \midrule
         \multirow{7}{*}{Linknet} & \multirow{2}{*}{4} & LD30 & 0.7828 & \textbf{0.1902} \\
          &  & LD50 & \textbf{0.8109} & 0.1752 \\ \cmidrule{2-5}
          & 6 & LD30 & 0.7866 & 0.1779 \\ \cmidrule{2-5}
          & \multirow{2}{*}{8} & LD30 & 0.7795 & 0.0555 \\
          &  & LD50 & 0.7996 & 0.0612 \\ \cmidrule{2-5}
          & \multirow{2}{*}{Standard} & LD30 & 0.7774 & 0.1212 \\ 
          &  & LD50 & \textbf{0.7892} & \textbf{0.1240} \\
             \midrule
         \multirow{6}{*}{U-Net} & \multirow{2}{*}{4} & LD30 & 0.7965 & 0.0537 \\
          &  & LD50 & 0.7886 & 0.0621 \\ \cmidrule{2-5}
          & \multirow{2}{*}{8} & LD30 & 0.7944 & 0.1644 \\ 
          &  & LD50 & 0.7800 & 0.1428 \\ \cmidrule{2-5}
          & Standard & LD30 & \textbf{0.7952} & 0.1405 \\
          \bottomrule 

    \end{tabular}
    \caption{Recall and precision values of the predictions that detected relict landslides. Numbers in bold represent the highest results for the CNN framework type.}
    \label{tbl:Predic_recall_pre}
\end{table}

\subsection{Discussion}
\label{sec:discussion}

This study found significant limitations in the semi-automatic detection of relict landslides in vegetation-covered areas using multispectral images as input to the CNN semantic segmentation process. The lack of labeled data for training the CNNs and the spectral similarities of the land cover units are probably the main reasons that prevented the CNNs to achieve more accurate results. 

The efficiency of the CNN classification ability relies mainly on the amount of labeled data used as input in the training process \citep{Ronneberger2015}. Usually, labeled landslide data is directly obtained from inventory maps that shows all landslides that have occurred over a period of time, however the inventory map of the study area is outdated and incomplete. It was fisrt created in the late 1970s to map a major landslide event that occurred in 1967 \citep{fulfaro1976escorregamentos}, and there have been no relevant updates since \citep{Correa2017}. Thus, labeled landslide data is sparse for the study area which made it necessary to perform data augmentation to enable CNN training even if augmentation may skew the data and decrease CNN's accuracy. 

The semi-automatic detection of relict landslides was only possible in this project due to the presence of \textit{Gleichenella sp.} ferns that cover degraded areas, prevent forest restoration, and preserve the boundaries of landslides (Section~\ref{sec:gleichenella}). As shown in Fig.~\ref{fig:Gleichenella}, the relict landslides can be distinguished from surroundings by the light green color of \textit{Gleichenella sp.} while native vegetation has a dark green color in general. However, there are spectral similarities between \textit{Gleichenella sp.} ferns and other land cover features such as pasture, agricultural fields, or deforested areas. Thus, the spectral characteristics are not enough to enable accurate detection of relict landslides and the occurrence of misclassification is somewhat expected. In areas covered by dense vegetation, such as the Atlantic rainforest, landslide detection is usually performed using multispectral images acquired soon after the event. In those images, landslides are easily recognizable due to the removal of material and soil exposure \citep{Guzzetti1999,Du2007,Booth2009,Burns2010,Sameen2019}. In multispectral images, the soil has a brown-orange color that contrasts with the surroundings' green color of vegetation cover, making identification easier for the CNNs, but even in these cases errors and misclassifications are common \citep{Soares2022}.   

These findings are essential to demonstrate that the accuracy of CNNs for landslide detection is closely related to the technical specificities of input data, availability of labeled landslide data and the study area characteristics, instead of the CNN's semantic segmentation process itself. For the best of the authors' knowledge this is the first project that uses CBERS4A images for landslide detection. The mid to high spatial resolution (8 m) and 4 spectral bands (R, G, B and NIR) were insufficient to accurately differentiate relict landslides from surroundings. Despite the adequate results of recall for all combinations ($\geq$ 75\%) and that 18 of 42 predictions correctly identified relict landslides, the majority of the results were inaccurate. \\

Also, a new CNN framework was proposed in an attempt to enhance landslide detection accuracy. The results demonstrate that the predictions of the proposed framework are better than those of the standard framework, although the values of precision and recall were quite similar. The k-means clustering process used to create the cluster dataset for the pre-training step computed weights to fine-tune the CNN training process, enhancing its predictions ability. Another finding is that it was not possible to address an optimal correlation of parameters that maximizes the accuracy of landslide detection (Section~\ref{sec:correlation}).

Most landslide detection studies are in areas partially covered with vegetation and focus on detecting recent landslides using post-event multispectral images when the landslides are clearly visible in the landscape  \citep{Chen2018,Ghorbanzadeh2019,Ji2020,Li2021}. Few studies occur in areas covered with rainforest vegetation \citep{Sameen2019, Soares2022} or for detecting relict landslides \citep{Wang2020}. Also, many studies use a combination of topographic and spectral information as input for the CNNs, which achieved worse results than using only spectral bands as input \citep{Ghorbanzadeh2019,Ji2020,Sameen2019,Wang2020,Soares2022}.

Thus, there is a lack of investigation in areas of tropical environments where landslides are common, and the monitoring is impaired by vegetation growth. The present study findings demonstrated the complexity of performing semi-automatic detection of relict landslides in rainforest areas using only multispectral imagery and highlighted the importance of the input data specifications for the CNN models. It also showed the relevance of a continuous update of inventory maps that increase labeled landslide data which may improve CNN landslide detection accuracy.

This study has several specificities that prevent directly comparing the results with other works. However, many studies of landslide detection using CNNs also showed that both the input data and the study area are directly related to model accuracy \citep{Chen2018,Ghorbanzadeh2019,Wang2020,Yu2021,Soares2022}. The CBERS4A multispectral images are the newer and best free data available for the mapping and monitoring of the brazilian territory. Despite the great advance promoted in the area of remote sensing for Brazil, the CBERS4A images were inappropriate for the detection of relict landslides in the specific conditions of this project. Nonetheless, they may be very useful and yield better results if used for recent landslide detection with images taken soon after the events. 

Although the results were generally inaccurate and the limitations of using multispectral images for relict landslide detection were exposed, the proposed framework had interesting results mainly from landslide predictions that outperformed the standard framework results. Even with all the limitations addressed previously, the CNN proposed framework could correctly predict relict landslides more or less accurately in almost half of the combinations (41.7\%) and achieved recall values higher than 75\% for every combination. The low precision values ($\leq$ 20\%) are related to the high rate of false positive samples predicted by the CNNs, but this is not necessarily inappropriate since it is suitable for relict landslide detection studies to predict more landslides than occur in the ground truth (Section~\ref{sec:validation}).

\section{Conclusion}

This work proposed a new CNN framework for semi-automatic relict landslide detection using CBERS4A multispectral imagery, and the results were compared with a standard CNN framework. The proposed framework is an attempt to improve CNN image segmentation ability and consenquently its accuracy by using transfer learning. It is composed by a pre-training step that uses a dataset generated by a k-means algorithm clustering process as input. The weights computed in the pre-training step are later used to fine-tune the CNN training process. Two augmentation factors of 30 and 50 were used to increase the dataset since labeled landslide data are sparse for the study area, and three CNNs (U-Net, Linknet and FPN) were used for image segmentation. Six combinations of CNN and parameters were generated for the standard framework and 36 combinations for the proposed framework.

The validation indices showed that the results of recall were $\geq$ 75\% for every combination and the precision results were generally between 10\% and 20\%. For the standard framework the higher recall value was 79.53\% (Unet/LD30) and the higher precision was 15.83\% (Unet/LD50) while for the proposed framework the higher values were 81.09\% (Linknet/4/LD50) for recall and 20.47\% (FPN/8/LD30) for precision. 

Predictions of the landslides for every combination tested were performed in a test area to enhance the evaluation process by visual interpretation of the relict landslides. 18 of 42 total combinations were able to correctly identify relict landslide more or less accurately. Predictions from the proposed framework were more accurate but they still showed many misclassifications mainly because of the amount of false positive samples. FPN was the more conservative model with less FP samples while U-Net and Linknet were less conservative and predicted more FP samples. In landslide detection studies, less conservative predictions models are adequate, since it is expected to detect more landslides than occur in ground truth and, thus, allow the detection of new landslides.  

Therefore, despite the recall and precision results were very similar for both frameworks, the predictions of the proposed framework were more accurate for the detection of relict landslides in the study area. Also, the results of validation indices and landslide predictions demonstrate that there is not an optimal correlation of parameters (type of CNN, number of clusters, augmentation factor), which increases the difficulties for the detection of relict landslides since the higher recall/precision values are not necessarily the most accurate in predicting relict landslides. 

Further investigation is needed to improve the proposed framework's accuracy, and the authors expect better results in areas with more visible landslides or using this framework for the detection of recent landslides. To improve detection of relict landslides, it is expected that using high to very high spatial resolution multispectral images, or even hyperspectral images, may overcome the limitations found in this study. Also, it is worthy testing if using DTMs from VHR topographic data as a single band, or as an additional band to RGB input data, in the CNN training process could enhance the accuracy of the models for relict landslide detection since DTMs do not contain the vegetation cover.

\section{Acknowledgments}
The authors acknowledge the support provided by the Institute of Energy and Environment, the Institute of Geosciences and the Graduate Program in Mineral Resources and Hydrogeology (PPG-RMH). Acknowledgements are extended to the Guest Editors, the Editor-in-Chief and the anonymous reviewers for their criticism and suggestions, which have helped to improve this paper.

\section{Disclosure Statement}
The authors report there are no competing interests to declare.

\section{Funding details}
This study was supported by the Sao Paulo Research Foundation (FAPESP) under grants \#2016/06628-0, \#2019/26568-0 and by Brazil’s National Council of Scientific and Technological Development, CNPq grant \#311209/2021-1  to C.H.G. This study was financed in part by CAPES Brasil - Finance Code 001. 

\section{Supplementary Data}
The supplementary material presents all 15 predictions that correctly detected landslides with TP, FP and FN results.

\section{Data Availability Statement}
Data Availability Statement: The code used in this research is available at the following link:
\url{https://github.com/SPAMLab/data_sharing/tree/main/Relict_landslides_CNN_kmeans} (accessed on 15
December 2022).

\section{ORCID}

Guilherme Pereira Bento Garcia \url{https://orcid.org/0000-0003-1209-7842}

Carlhos Henrique Grohmann de Carvalho \url{https://orcid.org/0000-0001-5073-5572}

Lucas Pedrosa Soares \url{https://orcid.org/0000-0002-6980-597X}

Mateus Espadoto \url{https://orcid.org/0000-0002-1922-4309}


\end{document}